\documentclass[letterpaper, 10 pt, conference]{ieeeconf}  % Comment this line out if you need a4paper

\IEEEoverridecommandlockouts                              % This command is only needed if 
\usepackage{graphics} % for pdf, bitmapped graphics files
\usepackage{subfig} % for subcaption
\usepackage{wrapfig}

\usepackage{amsthm} % for definition, theorm and etc
\usepackage{amsmath,amssymb}
\usepackage{graphicx}
\usepackage{tabularx}
\usepackage{multirow, makecell}
\usepackage{diagbox}
\usepackage{slashbox}
\usepackage{rotating}
\usepackage{cite}

\usepackage{url}
\usepackage{bm}
\usepackage[table]{xcolor}
\usepackage{hyperref}
\usepackage{siunitx} % degree symbol
\usepackage{booktabs}
\usepackage{cleveref}
\usepackage{mathtools} % for := \coloneqq , and split equation with raisetag. ex means x-height
\usepackage{upgreek} % to use uptau

\let\labelindent\relax % not including labelindent (avoid error for overloading \proof)
\usepackage{enumitem} % using enumerate with ref roman option

\usepackage{lipsum}

\usepackage[ruled,vlined]{algorithm2e}
\usepackage[nolist, nohyperlinks]{acronym}
\acrodef{MPC}[MPC]{Model Predictive Control}
\acrodef{QP}[QP]{Quadratic Program}
\acrodef{CBF}[CBF]{Control Barrier Function}

\newcommand{\vx}{{\boldsymbol x}}
\newcommand{\vu}{{\boldsymbol u}}

\newcommand{\vz}{{\boldsymbol z}}

\newcommand{\va}{{\boldsymbol a}} % block action
\newcommand{\vo}{{\boldsymbol o}} % observation

\newcommand{\StateSpace}{\mathcal{X}}
\newcommand{\ControlSpace}{\mathcal{U}}
\newcommand{\RealSpace}{\mathbb{R}}

\newcommand{\NaturalNumber}{\mathbb{N}}

\newcommand{\calC}{\mathcal{C}} % invariant set

\newcommand{\calP}{\mathcal{P}} % path
\newcommand{\calO}{\mathcal{O}} % obstacle
\newcommand{\calS}{\mathcal{S}} % safe set
\newcommand{\calN}{\mathcal{N}} % noraml distribution
\newcommand{\calL}{\mathcal{L}} % loss function
\newcommand{\calF}{\mathcal{F}}

\newcommand{\VSigma}{{\bm{\Sigma}}}
\newcommand{\vmu}{{\bm{\mu}}}

\newcommand{\horizon}{H} % MPC horizon

\DeclareMathOperator*{\argmin}{arg\,min}

\DeclareMathAlphabet{\mathmybb}{U}{bbold}{m}{n}
\newtheorem{proposition}{Proposition}
\theoremstyle{definition}
\theoremstyle{definition}

\theoremstyle{definition}

\theoremstyle{definition}

\theoremstyle{definition}

\usepackage{fontawesome5}
\newcommand{\shield}{\text{\tiny{\faShield*}}}

\newif\ifstatementboxes
\statementboxestrue             % ON; use \statementboxesfalse for OFF

\ifstatementboxes
  \usepackage[skins,breakable]{tcolorbox}
  \definecolor{statementblue}{RGB}{239,247,252}
  \definecolor{statementline}{RGB}{210,220,235}
  \definecolor{proofblue}{RGB}{251,253,255}
  \definecolor{proofline}{RGB}{238,242,247}

  \tcbset{
    paperbox/.style={
      enhanced, breakable,
      boxrule=0pt, frame hidden,
      arc=0pt, outer arc=0pt, boxsep=0pt,
      left=7pt, right=6pt, top=6pt, bottom=6pt,
      before skip=6pt, after skip=6pt
    },
    paperstatement/.style={
      paperbox,
      colback=statementblue, colframe=statementblue,
      borderline west={3pt}{0pt}{statementline}
    },
    paperproof/.style={
      paperbox,
      colback=proofblue, colframe=proofblue,
      borderline west={3pt}{0pt}{proofline}
    }
  }

  \tcolorboxenvironment{theorem}{paperstatement}
  \tcolorboxenvironment{lemma}{paperstatement}
  \tcolorboxenvironment{proposition}{paperstatement}
  \tcolorboxenvironment{proof}{paperproof}
\fi
\makeatother
\DeclareCaptionFont{mysize}{\fontsize{8}{9.6}\selectfont}
\title{\LARGE \bf JEPA-Shield: Safe Diffusion Planning with Latent World Models \\ via Conformal Shielding} 

\author{Taekyung Kim$^{1}$, Ege Yuceel$^{2}$, Steven Hong$^{3}$, Georgios Fainekos$^{3}$, Dimitra Panagou$^{1,4}$, Bardh Hoxha$^{3}$ % <-this % stops a space
\thanks{$^{1}$Department of Robotics, $^{4}$Department of Aerospace Engineering, University of Michigan, Ann Arbor, MI 48109, USA {\tt\footnotesize \{taekyung, dpanagou\}@umich.edu} } 
\thanks{$^{2}$Department of Electrical and Computer Engineering, University of Illinois Urbana-Champaign, Urbana, IL 61801, USA {\tt\footnotesize eyceel2@illinois.edu} } 
\thanks{$^{3}$Toyota Motor North America, Research \& Development, Ann Arbor, MI, 48105, USA {\tt\footnotesize <first\_name.last\_name>@toyota.com} }
}

\begin{document}
\maketitle
\thispagestyle{empty}
\pagestyle{empty}

\newcommand{\unsafeDecoder}{D}
\newcommand{\motionProbe}{Q}
\newcommand{\latentDynamics}{F}
\newcommand{\imageEncoder}{E}
\newcommand{\actionEncoder}{a_{\eta}}
\newcommand{\wrap}{\operatorname{wrap}}
\newcommand{\rel}{\operatorname{Rel}}
\newcommand{\Last}{\operatorname{Last}}
\newcommand{\Valid}{\operatorname{Valid}}
\newcommand{\Shield}{\operatorname{Shield}^{\shield}}
\newcommand{\dist}{\operatorname{dist}}

\newcommand{\fixed}[1]{\textcolor{black}{#1}}

\begin{abstract}
Latent world models enable planning from images, but their predictions do not directly establish whether a robot will satisfy physical constraints. We present JEPA-Shield, which connects a frozen action-conditioned Joint-Embedding Predictive Architecture (JEPA) world model to task-space safety enforcement. Two downstream models predict relative motion from imagined latent rollouts and decode unsafe geometry from the current image. Split conformal prediction turns end-to-end motion-prediction errors into multi-horizon uncertainty tubes. A backup-based shield checks these tubes against physical constraints and screens both candidate trajectories and final proposals in diffusion-inspired sampling-based model predictive control. Under the stated exchangeability and initial-backup assumptions, we establish probabilistic constraint satisfaction at action-block boundaries for a returned plan executed without replanning. We also introduce VParkBench, with geometric car and tractor-trailer parking tasks and a editable photorealistic Gaussian-splatting environment. In closed-loop evaluation, JEPA-Shield achieves success rates of 84.4\%, 82.3\%, and 99.0\%, respectively, with no observed constraint violations. Ablations show complementary benefits from shielding and uncertainty calibration. The results demonstrate the importance of grounding latent world-model predictions in task-space quantities and explicitly accounting for prediction uncertainty in safety-critical planning.
\href{https://www.taekyung.me/jepa-shield}{\textcolor{red}{[Project Page]}}\footnote{Project page: \href{https://www.taekyung.me/jepa-shield}{https://www.taekyung.me/jepa-shield}} \href{https://youtu.be/sN2jkkAWP8I}{\textcolor{red}{[Video]}} 
\end{abstract}

% \href{https://app.notion.com/p/JEPA-Shield-3dc2d61ec18b80d1b2a1d7806f3e4cab}{\textcolor{red}{[Project Page]}}\footnote{Anonymized project page: \href{https://app.notion.com/p/JEPA-Shield-3dc2d61ec18b80d1b2a1d7806f3e4cab}{https://app.notion.com/p/JEPA-Shield-3dc2d61ec18b80d1b2a1d7806f3e4cab}} 

%%% For the arXiv version, use these instead

% \href{https://www.taekyung.me/jepa-shield}{\textcolor{red}{[Project Page]}}\footnote{Project page: \href{https://www.taekyung.me/jepa-shield}{https://www.taekyung.me/jepa-shield}} \href{https://youtu.be/sN2jkkAWP8I}{\textcolor{red}{[Video]}} 

\section{INTRODUCTION}
\label{sec:introduction}

\begin{figure*}[t]
    \centering
    \includegraphics[width=\textwidth]{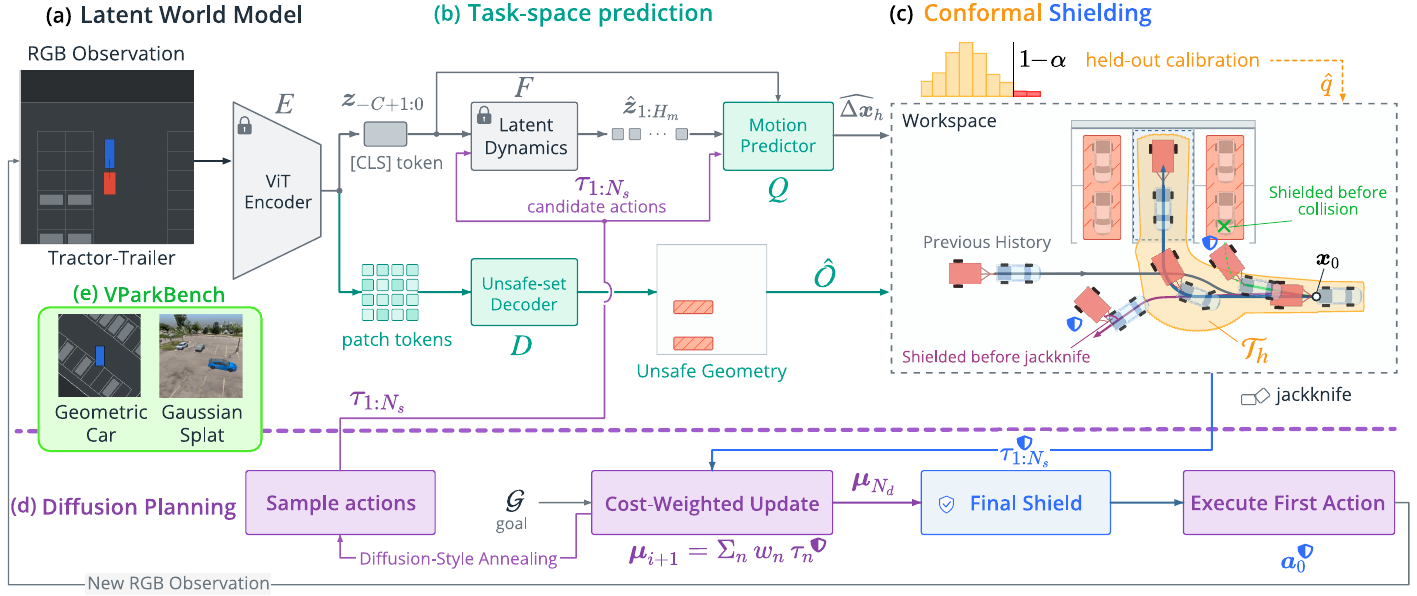}
    \caption{Overview of JEPA-Shield.
(a) A latent world model predicts action-conditioned future representations
from visual history.
(b) Task-space probes recover relative motion and decode unsafe geometry
from latent and spatial features.
(c) Conformal motion tubes enable uncertainty-aware shielding against
physical constraints.
(d) Diffusion planner iteratively updates shielded action
candidates and applies a final shield before execution.
(e) VParkBench evaluates the framework in geometric and
Gaussian-splatting parking environments.}
    \label{fig:block_diagram}
    \vspace{-5pt}
\end{figure*}

Planning from images requires more than predicting what a robot
will see next. A candidate maneuver must avoid obstacles, remain
within the workspace, and, for articulated vehicles, avoid jackknifing.
These requirements concern physical motion and geometry, not
similarity to a goal image. Optimization-based planning can enforce
explicitly modeled constraints, but errors in perception and scene
reconstruction can make their evaluation
unreliable~\cite{agrawal_certifiablycorrect_2025}.

World Models~(WMs) learn predictive representations from
observation-action trajectories, enabling image-based control through
imagined rollouts~\cite{ha_world_2018, hafner_dream_2020,hafner_mastering_2021}.
Joint-Embedding Predictive Architectures~(JEPAs) predict representations
rather than reconstructing pixels~\cite{assran_selfsupervised_2023}.
LeWorldModel~(Le-WM), for example, learns action-conditioned latent
dynamics through next-embedding prediction and
regularization~\cite{balestriero_lejepa_2025, maes_leworldmodel_2026},
then plans toward encoded visual goals. Such representations support
predictive control but need not expose the task-space quantities
\textbf{\emph{on which physical constraints are defined}}.

A downstream probe can predict safety values from latent rollouts,
but its outputs reflect errors in both the predicted latent trajectory
and the learned safety mapping. Near a
constraint boundary, an optimistic prediction can make an unsafe
maneuver appear acceptable. Thus, a favorable predicted safety
value alone does not establish physical constraint satisfaction.
Recent work has begun to address this gap. Kim \emph{et al.}
calibrate a learned binary failure predictor for model-predictive
shielding~\cite{kim_how_2026}; SLS$^2$ calibrates latent-dynamics and
constraint-prediction errors and propagates uncertainty sets through
robust model predictive control~(MPC)~\cite{nath_pixels_2026}; and
concurrent work by Zhong \emph{et al.} calibrates learned risk and
progress predictions before applying a separate deterministic
safety shield~\cite{zhong_calibrated_2026}.

We propose \textbf{JEPA-Shield} with a guiding principle:
\emph{predict in latent space, evaluate constraints in physical
space, and account for prediction error before accepting an action trajectory}.
A motion predictor converts the latent rollouts of a frozen JEPA
world model into multi-horizon relative trajectories, while a spatial
decoder estimates current unsafe geometry from encoder features.
Split conformal prediction calibrates the end-to-end motion error
to construct tubes for robust constraint evaluation.
A backup-based shield integrates these checks into diffusion-inspired
sampling-based MPC, retaining nominal prefixes only when their
backup-completed trajectories satisfy the robust constraints.
Under the stated initial-backup, exchangeability, and
conservative constraint-evaluation assumptions, the method provides
probabilistic constraint satisfaction at action-block endpoints
over a finite horizon.
Fig.~\ref{fig:block_diagram} summarizes the framework.

To evaluate the framework, we introduce \textbf{VParkBench},
a visual parking benchmark and dataset spanning geometric and
editable Gaussian-splatting environments.\footnote{The VParkBench dataset and generation code
will be released publicly pending completion of legal review.} On tractor-trailer parking,
JEPA-Shield increases success from 60.4\% to 82.3\% relative to
the unshielded ablation, while reducing the
observed unsafe rate from 34.4\% to 0\%.
Across all 288 benchmark trials, the full method achieves the
strongest overall performance with no observed constraint violations.

Our main contributions are:
\begin{itemize}[leftmargin=*]
    \item We introduce downstream models that recover multi-horizon
    relative motion from latent rollouts and decode current unsafe
    geometry from spatial features of a frozen JEPA world model.

    \item We calibrate end-to-end multi-horizon motion error to
    construct conformal tubes for robust physical constraint
    evaluation.

    \item We integrate conformal tubes with backup-based shielding
    in diffusion-inspired MPC, screening both sampled
    candidates and the final proposal before execution.

    \item We introduce \textbf{VParkBench}, a visual parking benchmark
    with geometric car and tractor-trailer tasks, an editable Gaussian-splatting environment, and collision, workspace,
    and jackknifing constraints.
\end{itemize}

\section{PRELIMINARIES}
\label{sec:preliminaries}

We first formulate constrained planning from visual histories and then summarize
the Le-WM and sampling-based diffusion planner components used by JEPA-Shield.

\subsection{Problem Formulation}
\label{subsec:problem_formulation}

Consider a discrete-time robotic system
\begin{equation}
\vx_{t+1}=f(\vx_t,\vu_t),
\label{eq:unknown_dynamics}
\end{equation}
where $t\in\NaturalNumber$ is the low-level time index,
$\vx_t\in\StateSpace\subseteq\RealSpace^n$ is the physical state,
$\vu_t\in\ControlSpace\subset\RealSpace^m$ is the control input, and
$f:\StateSpace\times\ControlSpace\rightarrow\StateSpace$ is unknown to the
online planner. The robot receives a visual observation

\begin{equation}
\vo_t=\operatorname{Obs}(\vx_t),\qquad
\operatorname{Obs}:\StateSpace\rightarrow
\RealSpace^{3\times N_{\textup{h}}\times N_{\textup{w}}},
\label{eq:general_visual_observation}
\end{equation}
where $N_{\textup{h}}$ and $N_{\textup{w}}$ are the image
height and width, respectively.
Physical-state labels are used for offline supervision and evaluation.
During online planning, future motion is predicted from the visual
context and candidate actions without querying $f$.

The world model may operate at a lower temporal resolution than the low-level
controller. Let $k\in\NaturalNumber$ denote the block-level time index. We
therefore group $r\in\NaturalNumber^{+}$ consecutive control inputs into one
block action, a practice also referred to as action chunking:
\begin{equation}
\va_k
\coloneqq
\begin{bmatrix}
\vu_{rk}^{\top} & \cdots & \vu_{rk+r-1}^{\top}
\end{bmatrix}^{\top}
\in
\mathcal{A}\subseteq\RealSpace^{rm},
\label{eq:general_action_block}
\end{equation}
where $\mathcal{A}\subseteq\ControlSpace^r$ is the admissible
block-action set. Hereafter, states and observations are indexed at block boundaries:
$\vx_k$ and $\vo_k$ refer to the state and observation at low-level
time $rk$. Thus, $\va_k$ is
applied between consecutive observations $\vo_k$ and $\vo_{k+1}$. At planning
cycle $k$, the
available observation-action context is
\begin{equation}
\mathcal{I}_k
\coloneqq
\left(
\vo_{k-C+1:k},
\va_{k-C+1:k-1}
\right),
\label{eq:general_real_context}
\end{equation}

where $C\in\NaturalNumber^{+}$ is the number of observations;
the $C-1$ action blocks connect consecutive observations.
Within each planning cycle, the current observation defines
local time zero. We write $\mathcal{I}=\mathcal{I}_k$ and
$\va_h=\va_{k+h\mid k}$ for the prediction index $h\in\NaturalNumber$;
state predictions and realized states use the same local
indexing. Thus, $\mathcal{I}=(\vo_{-C+1:0},\va_{-C+1:-1})$,
with negative indices denoting observed history. We restore
$k$ only when relating planning cycles. 
A candidate action sequence over a planning
horizon $\horizon\in\NaturalNumber^{+}$ is
\begin{equation}
\tau
\coloneqq
\left(
\va_0,
\ldots,
\va_{\horizon-1}
\right)
\in
\mathcal{A}^{\horizon}.
\label{eq:general_candidate_actions}
\end{equation}

Let $\Delta\vx_h=\rel(\vx_h;\vx_0)$ express the state in
the ego-centric frame of $\vo_0$, with angular differences
wrapped to their principal intervals. The relative-state
space is $\Delta\StateSpace\coloneqq
\{\rel(\vx;\vx_0):\vx\in\StateSpace\}\subseteq\RealSpace^n$,
with dependence on the fixed reference $\vx_0$ suppressed.
The learned rollout operator predicts
\begin{equation}
\widehat{\Delta\vx}_{1:\horizon}
\coloneqq
\widehat{\calF}(\mathcal{I},\tau)
\in(\Delta\StateSpace)^{\horizon}.
\label{eq:general_learned_rollout}
\end{equation}
Let $\widehat{\calO}$ denote an unsafe-set representation
inferred from the current visual context. The geometry
$\widehat{\calO}$, goal $\mathcal{G}$, and objective $J$ are
expressed in the same ego-centric frame. For
$N_g\in\NaturalNumber^{+}$ constraint functions $g_\ell$
on $\Delta\StateSpace$, the planning problem is
\begin{equation}
\begin{split}
\raisetag{8.0ex}
\tau^{\star}
& \in
\argmin_{\tau\in\mathcal{A}^{\horizon}}
\quad
J
\left(
\widehat{\Delta\vx}_{1:\horizon},
\tau;
\mathcal{G}
\right)\\
\textup{s.t.}\quad&
g_{\ell}
\left(
\widehat{\Delta\vx}_h,
\widehat{\calO}
\right)
\geq0, \, \ell=1,\ldots,N_g, \,
h=1,\ldots,\horizon.
\end{split}
\label{eq:general_constraints}
\end{equation}

The corresponding pointwise safe set is
\begin{equation}
\calC(\widehat{\calO})
\coloneqq
\left\{
\bar{\vx}\in\Delta\StateSpace
\ \middle|\
g_{\ell}(\bar{\vx},\widehat{\calO})\geq0,\,
\ell=1,\ldots,N_g
\right\}.
\label{eq:general_safe_set}
\end{equation}
The domain $\mathcal{A}^{\horizon}$ enforces the input bounds, and
$\widehat{\calF}$ determines the trajectory by forward rollout, as in
single shooting. Sec.~\ref{sec:method} instantiates $\widehat{\calF}$ and
$\widehat{\calO}$ using learned motion and unsafe-set models, and
incorporates prediction uncertainty into
\eqref{eq:general_constraints} through conformal shielding.

\subsection{LeWorldModel}
\label{subsec:lewm_preliminaries}

LeWorldModel~(Le-WM) is an action-conditioned JEPA world model that learns
predictive representations from offline observation-action
trajectories~\cite{maes_leworldmodel_2026}. Let $\imageEncoder$ denote its
vision-transformer~(ViT) observation encoder, $\latentDynamics$ its causal
action-conditioned latent dynamics, and $d_z$ the latent dimension. The
observation encoder produces
\begin{equation}
\vz_0=\imageEncoder(\vo_0)_{\textup{[CLS]}}, \label{eq:lewm_encoder_general}
\end{equation}
where $\vz_0\in\RealSpace^{d_z}$ denotes the class ([CLS]) token representation from
the final encoder layer~\cite{caron_emerging_2021}. For JEPA-Shield, we additionally retain the spatial
patch tokens produced by the same encoder for the unsafe-set decoder in
Sec.~\ref{subsec:unsafe_set_decoder}.

Given $C$ latent-action pairs, $\latentDynamics$
causally predicts the subsequent latent at each position
from the available history:
\begin{equation}
\widehat{\vz}_{-C+2:1}
=
\latentDynamics
\left(
\vz_{-C+1:0},
\va_{-C+1:0}
\right).
\label{eq:lewm_causal_prediction_general}
\end{equation}
During planning, encoding the observations in $\mathcal{I}$ gives
$\vz_{-C+1:0}$. The first candidate block $\va_0$ completes the action
input for predicting $\widehat{\vz}_1$. Longer imagined rollouts are
generated autoregressively: at each step, the predicted latent and
corresponding candidate action are appended to the fixed-length context
before $\latentDynamics$ is applied again.

Le-WM uses a mean-squared error~(MSE) next-embedding loss
$\calL_{\textup{pred}}$ and a stepwise SIGReg loss
$\calL_{\textup{SIG}}$ to prevent feature collapse. SIGReg regularizes
random one-dimensional embedding projections toward a standard Gaussian
distribution~\cite{balestriero_lejepa_2025, maes_leworldmodel_2026}.
The training objective is
\begin{equation}
\calL_{\textup{LeWM}}
\coloneqq
\calL_{\textup{pred}}
+
\lambda_{\textup{SIG}}
\calL_{\textup{SIG}},
\qquad
\lambda_{\textup{SIG}}\geq0.
\label{eq:lewm_total_loss_general}
\end{equation}

Since \eqref{eq:lewm_total_loss_general} contains no direct physical-state or
safety supervision, Sec.~\ref{sec:method} introduces a downstream motion
predictor and unsafe-set decoder to instantiate $\widehat{\calF}$ and
$\widehat{\calO}$, respectively.

%%%%%%%%%%%%%%%%%

\subsection{Diffusion-Inspired Sampling-Based MPC}
\label{subsec:dial_preliminaries}

Building on the rollout-based score estimation of Model-Based
Diffusion~\cite{pan_modelbased_2024}, DIAL-MPC~\cite{xue_fullorder_2025}
refines action sequences by sampling perturbations and
averaging candidates according to cost and constraint satisfaction,
with iteration- and horizon-dependent noise annealing.
For $\tau\in\mathcal{A}^{\horizon}$, let
$\mathcal{J}(\tau)\coloneqq
J(\widehat{\calF}(\mathcal{I},\tau),\tau;\mathcal{G})$
denote its rollout cost. The constrained target distribution
is~\cite{kim_safe_2026}
\begin{equation}
p_0(\tau)\propto p_J(\tau)p_g(\tau),\qquad
p_J(\tau)\propto e^{-\mathcal{J}(\tau)/\lambda},
\label{eq:constrained_target_distribution}
\end{equation}
where $\lambda>0$ is the temperature and $p_g(\tau)$
is the constraint-satisfaction indicator, equal to one when the predicted
rollout satisfies all inequalities in
\eqref{eq:general_constraints} and zero otherwise.

At iteration $i$, draw $N_{\textup{s}}$ candidates around the
stacked proposal mean $\vmu_i\in\RealSpace^{\horizon rm}$ and
update it by
\begin{equation}
\begin{aligned}
\tau_n
&=
\calP_{\mathcal{A}^{\horizon}}
\left(\vmu_i+\VSigma_i^{1/2}\boldsymbol{\epsilon}_n\right),
\quad
\boldsymbol{\epsilon}_n\sim
\calN(\boldsymbol{0},\boldsymbol{I}),\\
w_n
&=
\frac{p_0(\tau_n)}
{\sum_{j=1}^{N_{\textup{s}}}p_0(\tau_j)},
\qquad
\vmu_{i+1}
=
\sum_{n=1}^{N_{\textup{s}}}w_n\tau_n.
\end{aligned}
\label{eq:dial_update_general}
\end{equation}
Here, $n=1,\ldots,N_{\textup{s}}$,
$\VSigma_i$ is the perturbation covariance, and
$\calP_{\mathcal{A}^{\horizon}}$ projects onto the admissible
action-sequence set. Iteration indices are suppressed on samples
and weights.

Sparse feasible samples reduce sampling efficiency; if none are
feasible, the weights are undefined. Replacing hard feasibility
with finite constraint penalties does not guarantee constraint
satisfaction. Moreover, averaging feasible sequences can violate
the trajectory constraints~\cite{kim_safe_2026}.
These limitations motivate shielding both sampled candidates and
the final proposal in Sec.~\ref{sec:conformal_shielded_diffusion}. %%

%%%%%%%%%%%%%%%%%%%%%%%

\section{JEPA-SHIELD}
\label{sec:method}

JEPA-Shield separates visual prediction from physical
constraint evaluation. After training Le-WM with
\eqref{eq:lewm_total_loss_general}, we freeze its encoder
and latent dynamics. Two downstream models recover relative
motion from imagined rollouts and unsafe geometry from
current spatial features. Calibrated motion tubes and
backup-based shielding determine which proposed actions
can be retained with a valid backup.

\subsection{Multi-Horizon Relative Motion Prediction}
\label{subsec:motion_probe}

To recover robot motion from latent predictions, we train
a motion predictor $\motionProbe$ using physical-state
labels from the same offline dataset.
Let $H_{\textup{m}}\in\NaturalNumber^{+}$ denote the motion-prediction
horizon, with $H_{\textup{m}}\leq\horizon$.
Given the latent context $\vz_{-C+1:0}$ and the first
$H_{\textup{m}}$ action blocks of the candidate sequence $\tau$,
the frozen latent dynamics $\latentDynamics$ generates the imagined latents
$\widehat{\vz}_{1:H_{\textup{m}}}$. The motion predictor
$\motionProbe$ then predicts the relative-state sequence
$\widehat{\Delta\vx}_{1:H_{\textup{m}}}
\in(\Delta\StateSpace)^{H_{\textup{m}}}$ as
\begin{equation}
\widehat{\Delta\vx}_{1:H_{\textup{m}}}
=
\motionProbe\Big(
\vz_{-C+1:0},
\widehat{\vz}_{1:H_{\textup{m}}},
\va_{0:H_{\textup{m}}-1}
\Big).
\label{eq:motion_probe_operator}
\end{equation}

The motion predictor uses lightweight MLPs,
trained with a Huber loss at every prediction horizon. Its output at
horizon $h$ depends only on candidate actions $\va_{0:h-1}$. Only real
observations are encoded by $\imageEncoder$; future latents are generated
by the frozen predictor $\latentDynamics$. For
$\horizon>H_{\textup{m}}$, successive motion-prediction segments are
composed in the current real-observation frame. The autoregressive
construction, truncated at the requested horizon, instantiates
$\widehat{\calF}$ in~\eqref{eq:general_learned_rollout}. %%

%%%%%%%%%%%%%%

\subsection{Spatial Unsafe-Set Decoding}
\label{subsec:unsafe_set_decoder}

To evaluate predicted motion against the current scene,
we train an unsafe-set decoder $\unsafeDecoder$ from offline
binary labels. Following ViT-based dense
prediction~\cite{ranftl_vision_2021}, it maps the current
observation's spatial patch tokens to unsafe-set logits:
\begin{equation}
\widehat{\boldsymbol{M}}
=
D\!\left(E(\vo_0)_{\textup{patch}}\right)
\in
\RealSpace^{N_{\textup{h}}\times N_{\textup{w}}},
\label{eq:unsafe_decoder}
\end{equation}
and the corresponding unsafe probability map is
$\widehat{\boldsymbol{P}}
=
\operatorname{sigmoid}(\widehat{\boldsymbol{M}})$.

The decoder is trained against the binary unsafe-set mask
$\boldsymbol{M}\in\{0,1\}^{N_{\textup{h}}\times N_{\textup{w}}}$ using a
class- and boundary-weighted binary cross-entropy loss together with a
Tversky loss~\cite{salehi_tversky_2017}. We use asymmetric Tversky weighting
to penalize false negatives more strongly than false positives.

For a threshold $\tau_{\textup{occ}}\in(0,1)$, the decoded unsafe set is
\begin{equation}
\widehat{\calO}
\coloneqq
\left\{
\boldsymbol{\xi}\in
\{1,...,N_{\textup{h}}\}\times\{1,...,N_{\textup{w}}\}
:
[\widehat{\boldsymbol P}]_{\boldsymbol{\xi}}
\geq\tau_{\textup{occ}}
\right\},
\label{eq:decoded_unsafe_set}
\end{equation}
where $\boldsymbol{\xi}$ indexes the observation-aligned spatial grid.
The decoder is evaluated once per planning cycle, and
$\widehat{\calO}$ is shared by all candidate rollouts. We assume the represented
external unsafe geometry is stationary over the evaluated horizon. %%

%%%%%%%%%%%%%%%%%%%%%%%%%%%%%%%%%
\subsection{Conformal Motion Tube}
\label{subsec:conformal_tube}

A nominal trajectory is not enough for safety evaluation:
prediction errors can change whether a constraint is satisfied.
We therefore use split conformal
prediction~\cite{angelopoulos_conformal_2023} to calibrate
the complete composed trajectory, rather than its individual
prediction stages. Each held-out calibration trajectory
contributes one score: its largest normalized prediction
error across pose coordinates and horizons. A conformal
quantile of these scores determines the motion-tube widths.

Each calibration sample $c=1,\ldots,N_{\textup{cal}}$ contains a real
context $\mathcal{I}_c$, as in~\eqref{eq:general_real_context}, one
full-horizon action sequence $\tau_c\in\mathcal{A}^{\horizon}$, and its
realized relative-state trajectory $\Delta\vx^{(c)}_{1:\horizon}$.
Applying $\widehat{\calF}(\mathcal{I}_c,\tau_c)$ gives the corresponding
composed prediction $\widehat{\Delta\vx}^{(c)}_{1:\horizon}$.
Following~\cite{stamouli_recursively_2024,yu_signal_2026},
we normalize errors in the $n_{\textup{p}}$ geometric pose
coordinates. For relative states $\bar{\vx}$ and $\bar{\vx}'$, let
$\rho_d(\bar{\vx},\bar{\vx}')$ measure the
absolute difference in pose coordinate $d$, using wrapped
differences for scalar headings. Let $\sigma_{h,d}>0$ be
the corresponding scale, computed from training residuals
and fixed before calibration. The
normalized nonconformity score of calibration sample $c$ is

\begin{equation}
R_c
\coloneqq
\max_{\substack{h=1,\ldots,\horizon\\
d=1,\ldots,n_{\textup{p}}}}
\frac{
\rho_d\left(
\Delta\vx^{(c)}_h,
\widehat{\Delta\vx}^{(c)}_h
\right)
}{
\sigma_{h,d}
}.
\label{eq:joint_cp_score}
\end{equation}
The normalization prevents errors at horizons or coordinates with larger
numerical scales from dominating the maximum.

For the remainder of the construction, we fix the miscoverage level
$\alpha\in(0,1)$ and suppress dependence on it in the notation.
Define the conformal quantile 
\begin{equation}
\widehat q \coloneqq \operatorname{Quantile}_{1-\alpha} \left( R_1,\ldots,R_{N_{\textup{cal}}},+\infty \right),
\label{eq:cp_quantile}
\end{equation}
where $\operatorname{Quantile}_{1-\alpha}$ denotes the $\left\lceil(N_{\textup{cal}}+1)(1-\alpha)\right\rceil$-th smallest element of its arguments. %%

For a candidate sequence $\tau$, let $\widehat{\Delta\vx}_h(\tau)$ denote the $h$-th relative-state output of the complete rollout $\widehat{\calF}(\mathcal{I},\tau)$. Its conformal motion tube at horizon $h$ is
\begin{equation}
\mathcal{T}_h(\tau)
\coloneqq
\left\{
\bar{\vx}\in\Delta\StateSpace
\ \middle|\
\begin{array}{l}
\rho_d\!\left(
\bar{\vx},
\widehat{\Delta\vx}_h(\tau)
\right)
\leq\widehat q\,\sigma_{h,d},\\
\qquad d=1,\ldots,n_{\textup{p}}
\end{array}
\right\},
\label{eq:conformal_motion_tube}
\end{equation}
where $\bar{\vx}$ denotes a possible relative state.
The tube contains states consistent with the calibrated
motion error. Only the uncertain pose coordinates are
inflated; coordinates determined exactly by $\mathcal{I}$
and $\tau$ retain their horizon-$h$ values with zero
uncertainty width. For an articulated vehicle, orientation
uncertainty can substantially enlarge the possible trailer
footprint even when the nominal trajectory is nearly straight.

\begin{proposition}[Full-horizon motion coverage]
\label{prop:joint_cp_coverage}
Suppose that the learned rollout operator $\widehat{\calF}$ and
normalizing factors $\sigma_{h,d}$ are fixed independently of the
calibration samples, and that the $N_{\textup{cal}}$ calibration
samples and the next query, including their realized trajectories,
are exchangeable under the same fixed query-generation procedure. Then
\begin{equation}
    \mathbb{P} \left( \Delta\vx_h \in \mathcal{T}_h, \quad \forall h=1,\ldots,\horizon \right) \geq 1-\alpha.
\label{eq:joint_cp_coverage}
\end{equation}
\end{proposition}

\begin{proof}
Let $R_{\mathrm{new}}$ denote the normalized nonconformity score of the next query. Split conformal prediction gives $R_{\mathrm{new}}\leq\widehat q$ with probability at least $1-\alpha$. By the maximum in \eqref{eq:joint_cp_score}, this event implies the coordinate-wise tube bounds at every horizon simultaneously.
\end{proof}

The guarantee is marginal over exchangeable calibration and query
trajectories, not simultaneous over all candidates examined by the planner;
applying it to the retained plan therefore additionally requires that the
selected query satisfy the exchangeability assumption. For such a query,
the coverage event holds jointly across the prediction horizon, so robust
constraint evaluation over $\mathcal{T}_h$ also covers the realized state
at every horizon.

For each constraint $\ell$ and horizon $h$, define its robust value over the conformal motion tube as
\begin{equation}
\underline g_{\ell,h}(\tau)
\coloneqq
\inf_{\bar{\vx}\in\mathcal{T}_h(\tau)}
g_{\ell}\left(\bar{\vx},\widehat{\calO}\right).
\label{eq:robust_tube_constraint}
\end{equation}
For a single fixed query, we write $\underline g_{\ell,h}$.
The candidate trajectory is \textbf{\emph{robustly safe}} with respect to
$\widehat{\calO}$ when
$\underline g_{\ell,h}\geq0$
for every $\ell=1,\ldots,N_g$ and $h=1,\ldots,\horizon$.

%%%%%%%%%%%%%%%%%%%%%

\subsection{Diffusion Planning with Conformal Shielding}
\label{sec:conformal_shielded_diffusion}

We adapt the shielded rollout mechanism of Safe Model Predictive Diffusion
(Safe MPD)~\cite{kim_safe_2026} to the robust constraints in
\eqref{eq:robust_tube_constraint}. The shield modifies each candidate during
DIAL-MPC annealing and is applied again to the final proposal, since
weighted averaging does not preserve constraint satisfaction.
   
Let $H_{\textup{b}}\in\NaturalNumber^{+}$ be the backup horizon,
$\pi_{\textup{b}}$ a backup policy generating admissible action blocks, and
$\calS_0\subseteq\calC(\widehat{\calO})$ a terminal set that is forward
invariant under $\pi_{\textup{b}}$. 
Nominal rollouts from the real context $\mathcal{I}$ are
calibrated over $\horizon$ blocks. For the stopping backup used in this paper, tubes are extended analytically
through $\horizon+H_{\textup{b}}$, without predicting or
calibrating an additional backup tail.

For a nominal prefix of length $h\in\{0,\ldots,\horizon\}$, let
$\va^{\textup{b}}_{h:\horizon+H_{\textup{b}}-1}$ denote the backup
continuation generated by $\pi_{\textup{b}}$ from the predicted history, and define $\tau^{(h)} \coloneqq
\left(
\va_{0:h-1},
\va^{\textup{b}}_{h:\horizon+H_{\textup{b}}-1}
\right)$.
The first $H_{\textup{b}}$ backup blocks drive the predicted state toward
$\calS_0$, after which $\pi_{\textup{b}}$ is continued to complete the extended
horizon. For $h=0$, $\tau^{(0)}$ consists entirely of backup actions.
The sequence is valid if
\begin{equation}
\underline g_{\ell,j}\!\left(\tau^{(h)}\right)\geq 0,\quad
\forall \ell, j, 
\qquad
\mathcal{T}_{h+H_{\textup{b}}}\!\left(\tau^{(h)}\right)
\subseteq\calS_0,
\label{eq:prefix_backup_validity}
\end{equation}
where $\ell=1,\ldots,N_g$ and $j=1,\ldots,\max\{\horizon,h+H_{\textup{b}}\}$.

Starting from a valid $\tau^{(0)}$, the shield tests
$h=1,\ldots,\horizon$ sequentially and retains the latest valid completed
sequence. Testing stops at the first rejection or after all prefixes pass.
The first $\horizon$ blocks of the retained sequence form the shielded
sequence. Rejection therefore falls back to the previously validated
backup, not the backup generated from the rejected prefix. For fixed
$\mathcal{I}$, $\widehat{\calO}$, $\calS_0$, $\pi_{\textup{b}}$, $\widehat q$,
$\sigma_{h,d}$, and initial backup, denote this mapping by
$\operatorname{Shield}^{\shield}(\tau)$.

Algorithm~\ref{alg:jepa_shield} incorporates
$\operatorname{Shield}^{\shield}$ into $N_{\textup{d}}$ DIAL-MPC
annealing iterations. Every shielded candidate satisfies the robust
constraints, so $p_g$ in
\eqref{eq:constrained_target_distribution} is constant across these
samples and the weights depend only on $p_J$, avoiding the sampling inefficiency caused by infeasible samples~\cite{kim_safe_2026}. The final weighted
mean is shielded once more before execution, and the corresponding backup is retained for the next planning cycle.

\begin{figure*}[t]
    \centering
    \includegraphics[width=\textwidth]{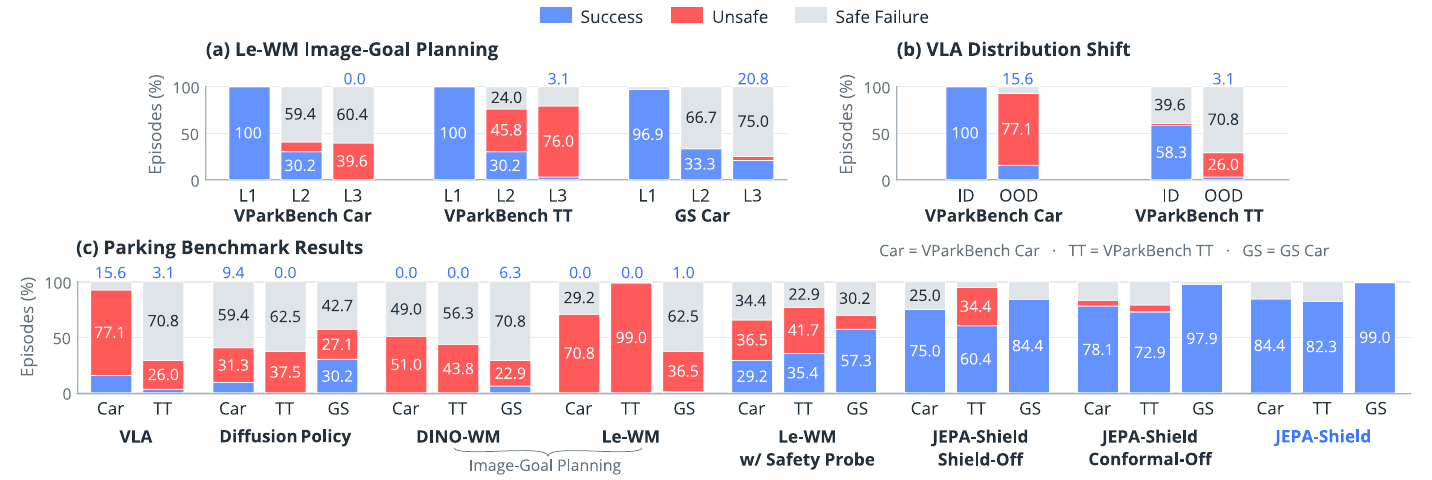}
    \caption{Quantitative parking evaluation, with 96 episodes per bar.
    (a) Le-WM CEM image-goal planning at increasing physical goal ranges:
L1 (1.0--1.2~m), L2 (2.4~m), and L3 (4.0~m). (b) VLA performance on exact training-episode replay (ID) versus
    unseen VParkBench scenarios (OOD). (c) Full parking benchmark comparison. Bars partition episodes into success, unsafe, and safe failure.
    Panel (a) uses latent-goal success; (b,c) use physical parking criteria.}
    \label{fig:parking_results}
    \vspace{-5pt}
\end{figure*}

\begin{algorithm}[h] \footnotesize
\caption{JEPA-Shield}
\DontPrintSemicolon
\label{alg:jepa_shield}
\KwIn{Context $\mathcal{I}$; goal $\mathcal{G}$; backup policy $\pi_{\textup{b}}$;
initial backup sequence $\tau^{(0)}$; terminal set $\calS_{0}$;
quantile $\widehat q$ and scales $\sigma_{h,d}$;
proposal $\vmu_0$}
\KwOut{Shielded sequence $\tau^{\shield}$ or failure}

Encode $\mathcal{I}$ and compute $\widehat{\calO}$
using~\eqref{eq:decoded_unsafe_set}\;

\If{$\tau^{(0)}$ fails \eqref{eq:prefix_backup_validity}}{
    \textbf{return} failure\;
}

\For{$i=0,\ldots,N_{\textup{d}}-1$}{
    \ForEach{$n=1,\ldots,N_{\textup{s}}$ \textup{in parallel}}{
        $\boldsymbol{\epsilon}_n
        \sim\calN(\boldsymbol{0},\boldsymbol{I})$\;

        $\tau_n\gets
        \calP_{\mathcal{A}^{\horizon}}
        \left(
        \vmu_i+\VSigma_i^{1/2}\boldsymbol{\epsilon}_n
        \right)$\;

        $\tau_n^{\shield} \gets
        \operatorname{Shield}^{\shield}(\tau_n)$\;
    }

    $\displaystyle
    w_n\gets
    \frac{p_J(\tau_n^{\shield})}
    {\sum_{j=1}^{N_{\textup{s}}}p_J(\tau_j^{\shield})},
    \quad n=1,\ldots,N_{\textup{s}}$\;

    $\displaystyle
    \vmu_{i+1}\gets
    \sum_{n=1}^{N_{\textup{s}}}w_n\tau_n^{\shield}$\;
}

$\tau^{\shield}\gets
\operatorname{Shield}^{\shield}
\left(\vmu_{N_{\textup{d}}}\right)$\;

\textbf{return} $\tau^{\shield}$\;
\vspace{-5pt}
\end{algorithm}

\begin{proposition}[Discrete-time constraint satisfaction]
\label{prop:discrete_safety}
Suppose the supplied initial backup satisfies
\eqref{eq:prefix_backup_validity}, and each
$g_\ell(\cdot,\widehat{\calO})$ conservatively lower-bounds its
corresponding physical constraint function. If the assumptions of
Proposition~\ref{prop:joint_cp_coverage} hold for the final retained
completed sequence at the extended calibration horizon, then executing the returned sequence without replanning satisfies the
physical constraints at local block times $1,\ldots,\horizon$ with probability at
least $1-\alpha$.
\end{proposition}

\begin{proof}
The final retained completed sequence satisfies the robust constraints at
every returned block time by construction.
Proposition~\ref{prop:joint_cp_coverage} gives simultaneous motion coverage
for this sequence. On this event,
\eqref{eq:robust_tube_constraint} implies physical constraint satisfaction
at all local block times $1,\ldots,\horizon$.
\end{proof}

Proposition~\ref{prop:discrete_safety} establishes discrete-time safety
for the certified plan and backup under the stated assumptions. Conformal
calibration covers motion-prediction error only; guarantees with decoded
geometry additionally require conservative constraint evaluation, which
can be incorporated independently and is outside the scope of this work.
The sampling-based update requires neither differentiation through the
learned models nor a learned denoising network. At planning cycle $k$,
Algorithm~\ref{alg:jepa_shield} uses $\mathcal{I}_k$ and, upon success,
executes only the first block of $\tau^{\shield}$ over the next $r$
low-level steps. After receiving $\vo_{k+1}$, the context and unsafe-set
estimate are updated. The next cycle is warm-started by shifting the
returned sequence by one block and appending the next block from its
retained backup continuation.

\begin{figure}[t]
    \centering

    \includegraphics[width=\linewidth]{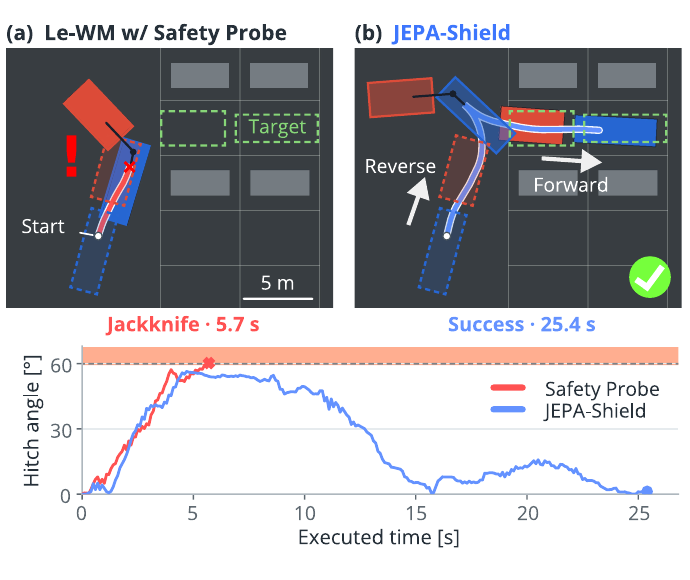}%
    \caption{Representative TT safety case.
    (a) Le-WM w/ Safety Probe enters a jackknife configuration after 5.7~s.
    (b) JEPA-Shield instead executes a reverse--forward maneuver and
    successfully parks in 25.4~s. Bottom: executed hitch-angle trajectories;
    the shaded region denotes the jackknife-unsafe range.}
    \label{fig:safety_case}
    \vspace{-5pt}
\end{figure}

\section{EXPERIMENTAL EVALUATION}
\label{sec:experiments}

\subsection{Experimental Setup}
\label{subsec:experimental_setup}

\begin{figure*}[t]
    \centering
    \includegraphics[width=\textwidth]{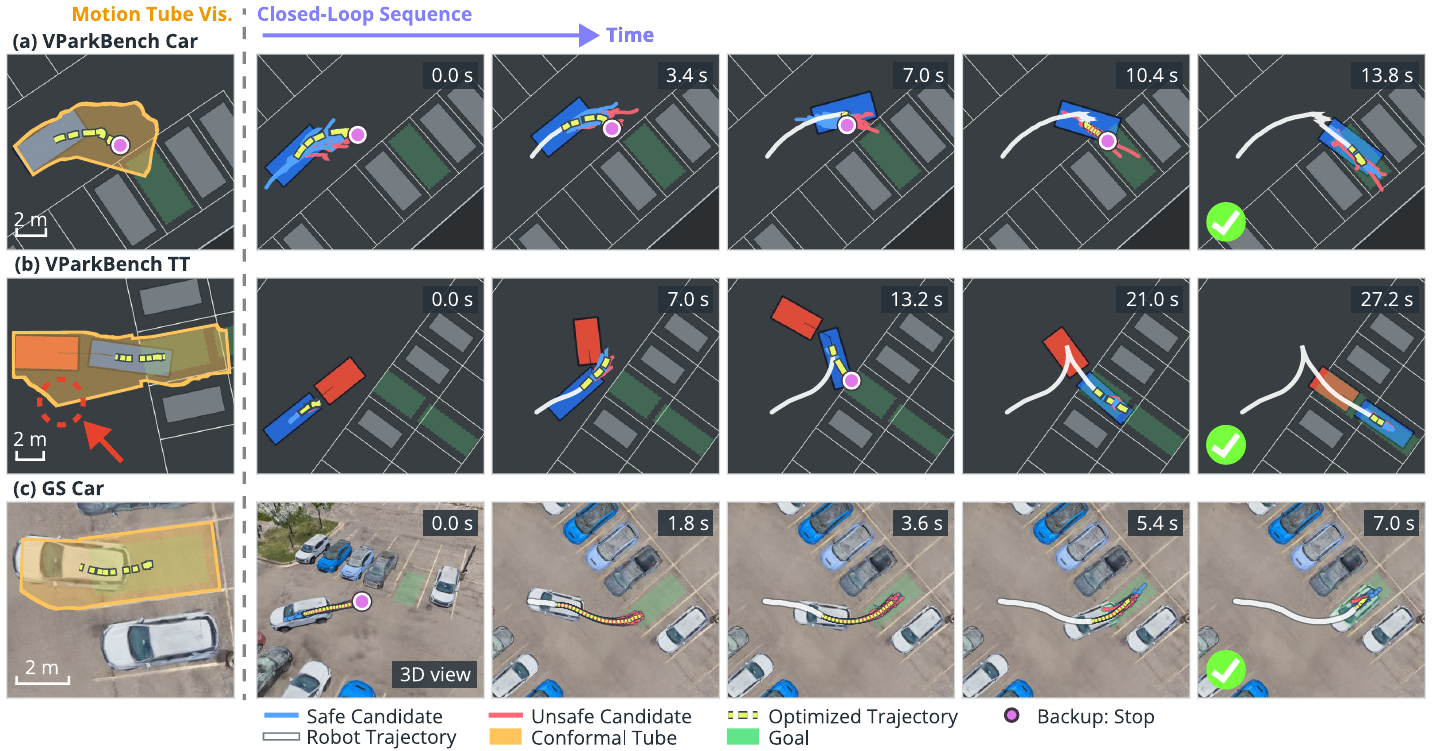}
    \caption{Qualitative JEPA-Shield results for
    (a) VParkBench Car, (b) VParkBench TT, and (c) GS Car.
    Left: conformal motion-tube visualization at one planning cycle.
    In (b), the red arrow highlights expansion of the yellow tube induced by
    hitch-angle uncertainty: although the nominal predicted motion is nearly
    straight, conformalization expands the possible articulated trailer placements,
    which are robustly checked by the shield.
    Right: closed-loop snapshots of complete successful parking maneuvers.}
    \label{fig:qualitative_results}
    \vspace{-5pt}
\end{figure*}

\subsubsection{VParkBench}

We introduce \textbf{VParkBench}, a visual parking benchmark for evaluating
goal-directed planning and control under safety constraints from bird's-eye RGB
observations (Fig.~\ref{fig:qualitative_results}). The benchmark includes
car (tractor-only) and tractor-trailer~(TT) parking tasks. The robot must reach a target parking configuration without collision
or workspace violations; TT must additionally avoid
jackknifing~\cite{kim_safe_2026}. Nonholonomic dynamics and articulated
geometry require maneuvering through highly nonconvex safe sets.

We represent the car and TT states as
\[
\begin{aligned}
\vx_{\textup{car}}
&=(x_c,y_c,\sin\theta,\cos\theta,v,\delta),\\
\vx_{\textup{TT}}
&=(x_r,y_r,
\sin\theta_{\textup{T}},\cos\theta_{\textup{T}},
\sin\theta_{\textup{R}},\cos\theta_{\textup{R}},
v,\delta).
\end{aligned}
\]
Here, $(x_c,y_c)$ and $(x_r,y_r)$ locate the car center and
tractor rear axle; $\theta$, $\theta_{\textup{T}}$, and
$\theta_{\textup{R}}$ are the car, tractor, and trailer headings;
and $v,\delta$ denote velocity and steering angle. Calibration uses $(x_c,y_c,\theta)$ for car and
$(x_r,y_r,\theta_{\textup{T}},\theta_{\textup{R}})$ for TT,
so $n_{\textup{p}}=3$ and $4$, respectively; $v,\delta$
are excluded from inflation. The commands $\vu=(v_{\textup{cmd}},\delta_{\textup{cmd}})$
are applied instantaneously at $0.10$-s control intervals,
with $|v_{\textup{cmd}}|\leq1.5$~m/s and steering limits of
$\pm35^\circ$ for car and $\pm45^\circ$ for TT.

The geometry-based simulator contains 16 parking spaces for car and 40 bays
arranged as 20 trailer pairs for TT. For the photorealistic
Gaussian-splatting~(GS) setting, we reconstruct a real 12-space parking
lot from video using G4Splat~\cite{ni_g4splat_2026}. Following
FalconGym~2.0's editable GS
framework~\cite{miao_performanceguided_2025}, we embed vehicle assets
in the scene and couple their poses to the vehicle dynamics.
The resulting \textbf{GS Car} environment has a smaller physical
operating domain than the geometric environments; its quantitative
evaluation covers car parking only. Each setting contains 96 scenarios: 48 forward and 48 reverse parking
tasks, with left-, right-, and both-side neighboring occupancy.
Every scenario requires substantial turning, and some require multiple
turns in tight spaces. Initial straight-line distances from the vehicle
body to the goal range from 6.8 to 11.8~m across the benchmark.

\subsubsection{Dataset, Training and Planning Details}
Offline trajectories contain bird's-eye RGB observations, controls,
physical states, and simulator-derived unsafe-set and minimum-clearance
labels. Training uses approximately 12.4~k, 12.4~k, and 13.0~k episodes
for geometric car, geometric TT, and GS Car, respectively. Validation
and calibration data are held out from model training. In addition, the
96 evaluation scenarios in each VParkBench setting are excluded from
training and contain unseen task-space goals and parking configurations. JEPA-Shield follows the training procedure in
Sec.~\ref{sec:method}. Learned baselines use the same task-specific data
unless stated otherwise; Diffusion Policy uses only the successful
expert subset.

We use $\alpha=0.05$ and $N_{\textup{cal}}=1000$.
Following~\cite{stamouli_recursively_2024,yu_signal_2026},
$\sigma_{h,d}$ is the maximum training discrepancy
$\rho_d$ at horizon $h$. One query per held-out trajectory
uses three observation frames and the subsequent recorded
expert/controller actions, with predictions recursively
composed over $\horizon$ blocks.
All environments use
$(r,C,H_{\textup{m}},H_{\textup{b}})=(2,3,8,6)$,
with $0.20$-s action blocks. For planning cycles after initialization,
$(\horizon,N_{\textup{s}},N_{\textup{d}})$ are
$(32,192,7)$, $(24,640,6)$, and $(24,160,6)$ for geometric car, geometric TT, and GS Car, respectively.
We set $\tau_{\textup{occ}}=0.5$. Following Safe MPD~\cite{kim_safe_2026}, we use a stopping backup policy described in
Sec.~\ref{sec:conformal_shielded_diffusion}, with
$\calS_0=\{\vx\in\calC(\widehat{\calO}):v=0\}$.

\subsubsection{Baselines and Ablations}
We compare against both policy-based and world-model-based baselines.
\emph{Diffusion Policy}~\cite{chi_diffusion_2024} directly predicts action
sequences from RGB observations. \emph{VLA} denotes an NVIDIA GR00T~N1.7~3B
policy~\cite{nvidia_gr00t_short} separately fine-tuned for the VParkBench car
and TT tasks following NVIDIA's official custom-embodiment
(\texttt{NEW\_EMBODIMENT}) fine-tuning workflow. It receives RGB observations and a state/relative-goal
representation and predicts control blocks.
\emph{DINO-WM}~\cite{zhou_dinowm_2025} and
\emph{Le-WM}~\cite{maes_leworldmodel_2026} use their original
cross-entropy method~(CEM) planning procedures to minimize latent
distance to an encoded goal observation. We provide the required goal
images, although these are not generally available for arbitrary unseen
destinations.

We further consider \emph{Le-WM w/ Safety Probe}, which augments Le-WM with
a downstream predictor of minimum obstacle clearance. The predicted clearance
is evaluated along sampled future rollouts, and candidates with negative
predicted clearance receive a large safety penalty in our diffusion planner's objective.
The task-space world-model planners, including JEPA-Shield, use an ego-relative
goal representation; for TT, the current hitch angle is additionally provided,
as this quantity is readily measurable with inexpensive onboard sensing. Their task costs use
the same position, heading, and control terms, and all penalty-based safety
terms use the same penalty weight.

Finally, we evaluate two ablations. \emph{JEPA-Shield-Off} removes
$\operatorname{Shield}^{\shield}$ while retaining the learned task-space
prediction and planning objective. \emph{JEPA-Shield-Conformal-Off}
retains the shield but removes conformal uncertainty
inflation, so constraint checks are performed only on the nominal
predicted trajectory. Full \emph{JEPA-Shield} uses both
conformal motion tubes and backup-based shielding.

\subsubsection{Evaluation Protocol}
An episode is classified as \emph{unsafe} if it contains a collision,
workspace violation, or, for TT, jackknifing. A violation-free episode
is successful if its final position error is at most $0.8$~m and its
heading error is at most $15^\circ$. For TT, both tractor and trailer
headings must satisfy the corresponding accepted goal configuration.
All other episodes are \emph{safe failures}. The separate image-goal
evaluation instead requires the observed latent MSE to fall below its
checkpoint-specific threshold and to decrease by at least 90\% from its
initial value; any unsafe transition precludes success.

\subsection{Results and Analysis}
\label{subsec:results_analysis}

We organize the evaluation around four questions:
\textbf{Q1}: How does image-goal planning scale with physical goal range?
\textbf{Q2}: How well do policies learned from demonstrations, including a fine-tuned VLA, generalize to unseen parking goals and configurations?
\textbf{Q3}: What are the contributions of task-space prediction,
shielding, and conformal uncertainty calibration?
\textbf{Q4}: How does performance change in the photorealistic GS setting?

\textbf{Image-goal planning (Q1).}
Fig.~\ref{fig:parking_results}(a) evaluates Le-WM CEM as the physical
goal range increases. At Level~1 (L1), success is 100.0\%, 100.0\%, and
96.9\% for VParkBench Car, VParkBench TT, and GS Car, respectively,
but falls to 30.2--33.3\% at L2 and 0.0--20.8\% at L3.
Although the levels use different held-out trajectories and therefore do
not isolate distance alone, the full parking benchmark in
Fig.~\ref{fig:parking_results}(c) shows the same limitation:
DINO-WM and Le-WM achieve no success on the geometric car or TT tasks
and only 6.3\% and 1.0\% on GS Car, respectively. 
These results suggest that latent image matching provides useful local
guidance but becomes less reliable in the evaluated longer-range tasks.
This is consistent with accumulating rollout error and the absence of
explicit task-space safety constraints in the latent goal objective.

\textbf{Policy generalization (Q2).}
Fig.~\ref{fig:parking_results}(b) shows a pronounced distribution shift
for the VLA. Car success decreases from 100.0\% on exact training-episode
replay to 15.6\% on unseen VParkBench scenarios, while TT success
decreases from 58.3\% to 3.1\%. The corresponding OOD car runs are
77.1\% unsafe, while 70.8\% of the TT runs terminate as safe failures.
Likewise, Diffusion Policy achieves only 9.4\%, 0.0\%, and 30.2\%
success on car, TT, and GS Car, respectively
(Fig.~\ref{fig:parking_results}(c)). 
Thus, policies learned from demonstrations, including fine-tuning a large
pretrained VLA, do not by themselves provide reliable generalization to
unseen parking configurations or safety constraints.

\textbf{Task-space safety and ablations (Q3).}
Even without shielding, the task-space JEPA planner provides a substantial
gain: JEPA-Shield-Off achieves 75.0\%, 60.4\%, and 84.4\% success on
car, TT, and GS Car, outperforming all non-JEPA-Shield baselines.
In comparison, Le-WM w/ Safety Probe reaches 29.2\%, 35.4\%, and
57.3\% success, while producing 36.5\%, 41.7\%, and 12.5\% unsafe
episodes. A learned scalar safety prediction evaluated along recursively
imagined rollouts remains subject to rollout and probe errors, and a
finite penalty does not constitute a constraint certificate.
Fig.~\ref{fig:safety_case} shows a representative TT comparison:
Le-WM w/ Safety Probe reaches a jackknife configuration, whereas
JEPA-Shield completes the maneuver while maintaining the articulation
constraint.

Full JEPA-Shield achieves 84.4\%, 82.3\%, and 99.0\% success with no
observed unsafe episodes across the 288 benchmark trials. Removing the
conformal expansion reduces success to 78.1\%, 72.9\%, and 97.9\% and
reintroduces unsafe episodes in the geometric settings. The mechanism is
illustrated in Fig.~\ref{fig:qualitative_results}(b): because the
conformal tube accounts for uncertainty in the predicted pose
coordinates, uncertainty in the hitch angle enlarges the possible trailer
occupancy even when the nominal trajectory is nearly straight. The shield therefore evaluates worst-case articulated placements within
the calibrated tube rather than only the nominal footprint. 
Without shielding, TT has 34.4\% unsafe episodes.
Together, the ablations demonstrate the complementary contributions of
shielding and conformal uncertainty calibration.

\textbf{Photorealistic visual setting (Q4).}
Overall, performance is generally higher on GS Car than on the geometric
car benchmark for methods evaluated in both environments. Thus, despite
the substantially more realistic visual appearance, visual complexity
alone does not appear to dominate performance in these experiments.
This comparison is not controlled for appearance alone: GS Car also has
a physically smaller operating domain and different maneuvering geometry.
Rather, the results suggest that planning performance depends strongly on
the task-space geometry and safe motion required to reach the goal, in addition
to visual representation quality. Fig.~\ref{fig:qualitative_results}(c)
shows a complete JEPA-Shield maneuver in the reconstructed GS scene.

\section{Conclusion}

JEPA-Shield connects latent world-model prediction with explicit evaluation of physical constraints. Its motion tubes account for prediction error, while its backup-based shield determines which proposed actions can be retained. Across geometric and photorealistic parking tasks, the framework combines strong task completion with no observed constraint violations in 288 trials. The central lesson is that useful visual predictions and reliable constraint evaluation are distinct requirements: a world model can supply the former, while task-space reasoning and uncertainty-aware shielding address the latter.

\addtolength{\textheight}{0 cm}   % This command serves to balance the column lengths
                                  % on the last page of the document manually. It shortens
                                  % the textheight of the last page by a suitable amount.
                                  % This command does not take effect until the next page
                                  % so it should come on the page before the last. Make
                                  % sure that you do not shorten the textheight too much.

%%%%%%%%%%%%%%%%%%%%%%%%%%%%%%%%%%%%%%%%%%%%%%%%%%%%%%%%%%%%%%%%%%%%%%%%%%%%%%%%

%%%%%%%%%%%%%%%%%%%%%%%%%%%%%%%%%%%%%%%%%%%%%%%%%%%%%%%%%%%%%%%%%%%%%%%%%%%%%%%%

%%%%%%%%%%%%%%%%%%%%%%%%%%%%%%%%%%%%%%%%%%%%%%%%%%%%%%%%%%%%%%%%%%%%%%%%%%%%%%%%

%%%%%%%%%%%%%%%%%%%%%%%%%%%%%%%%%%%%%%%%%%%%%%%%%%%%%%%%%%%%%%%%%%%%%%%%%%%%%%%%
\bibliographystyle{IEEEtran}
\typeout{}
\bibliography{references, local_refs}

\end{document}